\documentclass{article}
\usepackage{amsmath}

\usepackage{arxiv}

\usepackage{xcolor}%
\usepackage{algorithm}%
\usepackage{algpseudocode}%
\usepackage{listings}%
\usepackage{footnotebackref}

\usepackage[utf8]{inputenc} % allow utf-8 input
\usepackage[T1]{fontenc}    % use 8-bit T1 fonts
\usepackage{hyperref}       % hyperlinks
\usepackage{url}            % simple URL typesetting
\usepackage{booktabs}       % professional-quality tables
\usepackage{amsfonts}       % blackboard math symbols
\usepackage{nicefrac}       % compact symbols for 1/2, etc.
\usepackage{microtype}      % microtypography
\usepackage{lipsum}
\usepackage{graphicx}
\usepackage{cleveref}
\usepackage{multirow}
\usepackage{amsmath,amssymb,amsfonts}%
\usepackage{amsthm}%
\usepackage{mathrsfs}%
\usepackage{subfigure}
\usepackage{txfonts, bbding}
\usepackage{mathtools}
\usepackage{booktabs}%
\usepackage{amsthm}
\graphicspath{ {./images/} }

\newtheoremstyle{thmstyleone}% Name
  {18pt plus2pt minus1pt}  % Space above
  {18pt plus2pt minus1pt}  % Space below
  {\normalfont}      % Body font
  {0pt}                    % Indent amount
  {\itshape}         % Theorem head font
  {}                       % Punctuation after theorem head
  {0.5em}                  % Space after theorem head
  {\thmname{#1}~\thmnumber{#2}%
   \ifx#3\empty\else\thmnote{ {\normalfont(#3)}}\fi} % Properly handling optional notes
\theoremstyle{thmstyleone}%
\newtheorem{theorem}{Theorem}
\newtheorem{lemma}{Lemma}
\newtheorem{definition}{Definition}%

\title{Incentivizing Inclusive Contributions in Model Sharing Markets}

\author{
 Enpei Zhang\textsuperscript{1*} \\
   \And
 Jingyi Chai\textsuperscript{1*} \\
  \And
 Rui Ye\textsuperscript{1*} \\ 
 \And
 Yanfeng Wang\textsuperscript{1} \\
 \And
 Siheng Chen\textsuperscript{1} \\\\
 \textsuperscript{1} Shanghai Jiao Tong University \\
 * These authors contributed equally to this work.
}
\begin{document}
\maketitle

\begin{abstract}
While data plays a crucial role in training contemporary AI models, it is acknowledged that valuable public data will be exhausted in a few years, directing the world's attention towards the massive decentralized private data.
However, the privacy-sensitive nature of raw data and lack of incentive mechanism prevent these valuable data from being fully exploited.
Addressing these challenges, this paper proposes inclusive and incentivized personalized federated learning (iPFL), which incentivizes data holders with diverse purposes to collaboratively train personalized models without revealing raw data.
iPFL constructs a model-sharing market by solving a graph-based training optimization and incorporates an incentive mechanism based on game theory principles.
Theoretical analysis shows that iPFL adheres to two key incentive properties: individual rationality and truthfulness.
Empirical studies on eleven AI tasks (e.g., large language models' instruction-following tasks) demonstrate that iPFL consistently achieves the highest economic utility, and better or comparable model performance compared to baseline methods.
We anticipate that our iPFL can serve as a valuable technique for boosting future AI models on decentralized private data while making everyone satisfied.
\end{abstract}

% keywords can be removed
\section{Introduction}\label{sec1}

Training on massive publicly-available data~\cite{c4,gao2020pile,laion400m,laion5b}, AI models have demonstrated significant proficiency in diverse domains~\cite{gpt3,gpt3.5,latent-diffusion,dalle}.
As a well-known representative, ChatGPT~\cite{gpt3.5,gpt4} has swept the world with its exceptional ability to solve general tasks.
While it is commonly acknowledged that more data leads to better performance~\cite{kaplan2020scaling}, it has been estimated that available and valuable data in public will be exhausted by the year 2026~\cite{villalobos2022will,muennighoff2023scaling}, significantly impeding the continued enhancement of AI models under the current training paradigm.

The gradual depletion of public data starkly contrasts with the private sector, where massive institutions separately hold a wealth of valuable data.
For instance, financial institutions such as Bloomberg~\cite{wu2023bloomberggpt} possess high-quality private data to train AI models for finance.
Ideally, if these institutions collaborate on their resources, they can create a substantial and diverse database capable of augmenting contemporary AI models~\cite{wu2023bloomberggpt,med-palm,wang2023far,chen2023pointgpt}.
Unfortunately, two critical practical issues prevent distributed private data from being fully exploited~\cite {gdpr,kairouz2021advances}.
Firstly, the sensitivity of private data deters institutions from sharing it readily since this could raise privacy concerns and cause interest conflict~\cite{gdpr,price2019privacy,hathaliya2020exhaustive,box2013improving, qi2023differentially,kaissis2021end}.
Secondly, the absence of a comprehensive incentive mechanism results in a lack of motivation for institutions to actively and willingly engage in collaboration~\cite{yang-survey,karimireddy2022mechanisms}.

Consequently, to enable the utilization of decentralized private data for the continued enhancement of contemporary AI models, it is imperative to establish a harmonious sharing market, which should safeguard privacy and ensure individual interests.
In this market, data owners could act as buyers who selectively buy models from others to help train stronger models for their interested tasks; or as sellers who gain revenues from other institutions that have bought their models.
Such a guarantee of privacy (i.e., trading models rather than data) and interests can well motivate institutions to participate in the market, forming a virtuous circle as more participants lead to better performance which in turn attracts more participants.

\begin{figure}
    \centering
    \includegraphics[width=1\linewidth]{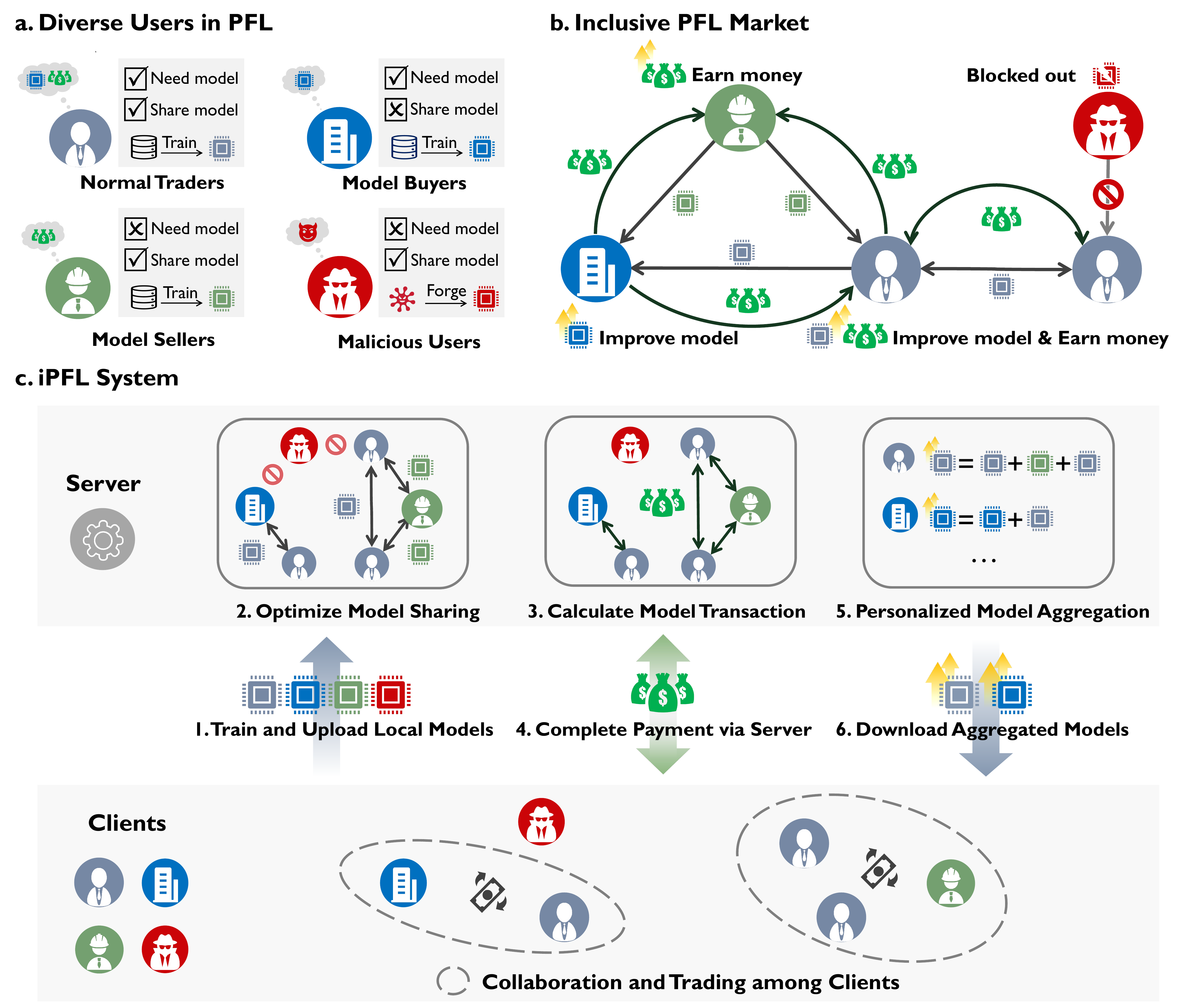}
    \caption{Inclusive PFL market and our iPFL. \textbf{a.} The clients have different purposes for entering a PFL system. A client can be: i) a trader who simultaneously buys model and sells their model; ii) a buyer who only buys a model and never shares its own model; iii) a seller who only sells its own model and never buys models; iv) an attacker who intends to ruin the system. \textbf{b.} In an inclusive market system, the model and money transaction should satisfy the needs of all the participants and block out attackers. \textbf{c.} In our iPFL, all the market behaviors are completed over a neutral server.}
    \label{fig:overview}
\end{figure}

Following this vision, we adopt personalized federated learning (PFL)~\cite{wu2022fedgraph,pfedme,fallah2020personalized} as the technical foundation for model training in this market, due to PFL's properties on preserving data privacy (i.e., sharing models) and catering to personal interests (i.e., improving personalization performance).
In this PFL-based market, coordinated by a central server, participants share their locally-trained models to achieve personalization through collaboration~\cite{pfedgraph,ditto,fedamp}. 
This approach has shown promising personalized performance through techniques like model regularization~\cite{ditto}, meta-learning~\cite{fallah2020personalized}, and clustering~\cite{cfl}.
However, existing methods mainly focus on personalization techniques, overlooking participants' economic conditions and motivations, which are two key factors in market dynamics.

Therefore, in this paper, we introduce an inclusive PFL system that accommodates individual model preferences and economic conditions, where we specifically consider four types of participants as shown in~\cref{fig:overview} (a).
We model the overall system as a graphical game, with participants as nodes and their exchange relationships as asymmetrically weighted edges, enabling a nuanced model-sharing network; see illustration in~\cref{fig:overview} (b).
To achieve this, we propose a novel graph-based PFL optimization objective that captures an individual's model preference via model similarity and economic conditions via reserving personalized utility functions.
Specifically, we pursue personalized models by minimizing loss on interested tasks while maximizing the pair-wise model similarity among participants and the total social welfare within the overall collaboration graph.
In this way, participants are allowed to select models based on their preferences and affordability, improving personalization performance, enhancing system robustness against inauthentic models and promoting cost efficiency.

While the graph-based PFL provides the technical foundation, the market's success also depends on an effective incentive mechanism to motivate participation.
This mechanism must fairly reward contributions and ensure those benefiting from contributions compensate accordingly, while also promoting honest participation and deterring dishonest or malicious behavior~\cite{li2021boosting,zhan2020learning}.
To achieve this, we design a payment mechanism in our PFL system (we term our overall system iPFL where i denotes incentivized and inclusive) that encourages willing and honest participation.
This mechanism sets specific prices for model transactions, calculated using game theory principles and considering both the buyer's economic utility and the seller's model quality.
This ensures mutual benefit from each transaction.
Through theoretical analysis, we show that iPFL adheres to two key incentive principles: individual rationality, ensuring that all participants benefit from each training round, and truthfulness, incentivizing clients to disclose their true training costs, fostering a collaborative and honest market environment.

To verify the effectiveness of our proposed iPFL (see system overview in~\cref{fig:overview}), we conduct extensive experiments, covering comprehensive comparisons with baselines, diverse scenarios and tasks.
Results show that iPFL consistently achieves higher economic utility, and better or comparable personalization performance compared to state-of-the-art PFL methods.
Remarkably, in a scenario of training large language models~\cite{llama2}, iPFL can achieve $49\%$ higher economic utility and $9\%$ higher model utility than the best baseline method.
We anticipate that our proposed iPFL can serve as a valuable technique for boosting future AI models on decentralized private data while making everyone satisfied.

\section{Results}\label{sec2}

% \subsection{Experimental Setups}
\begin{figure}
    \centering
    \includegraphics[width=.98\linewidth]{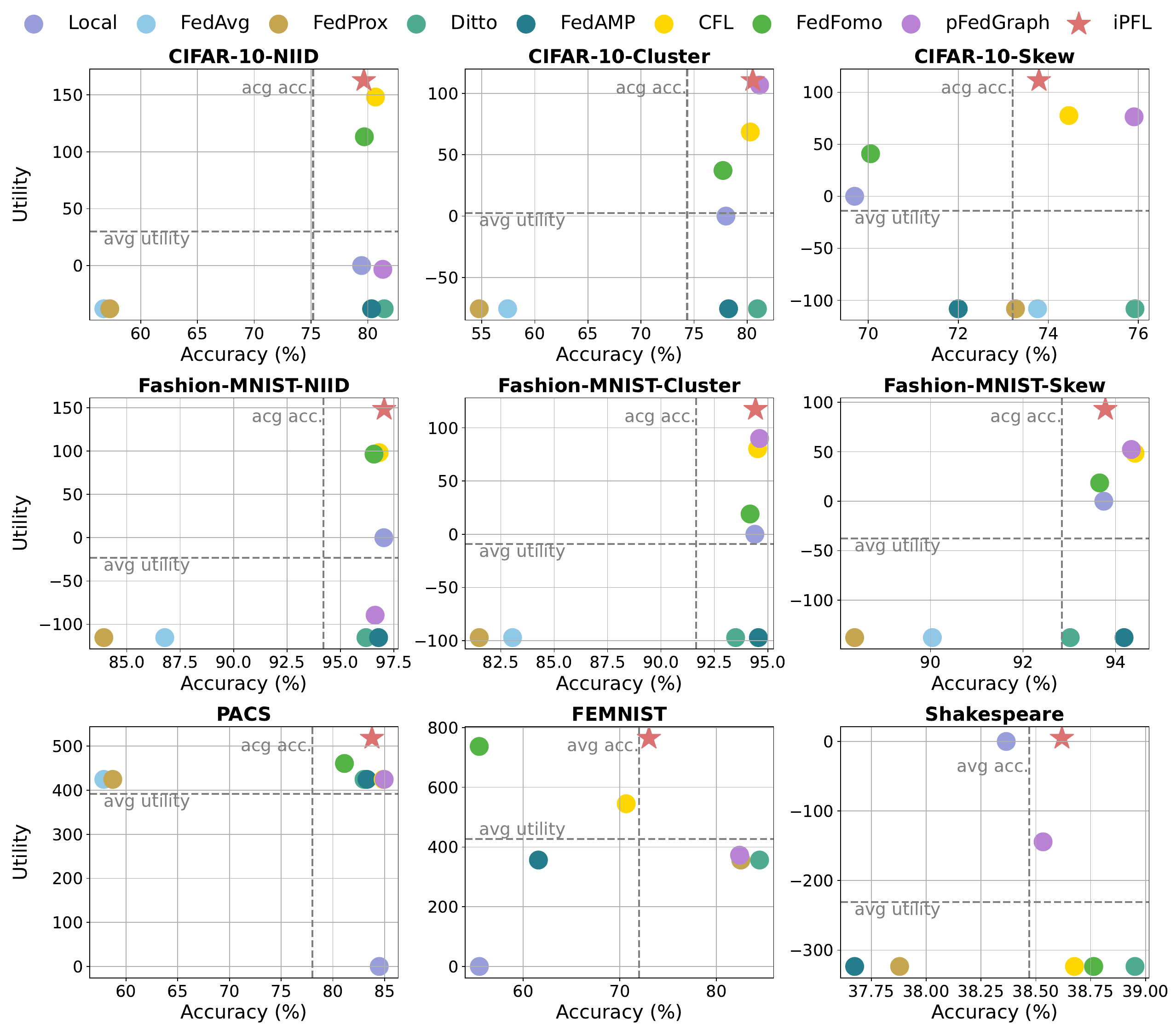}
    \caption{Comparison of average utility and accuracy in scatter under different settings. NIID represents the $\beta=0.1$, Cluster stands for 3 or 4 groups with $\beta=0.1$ among clusters, consider two cases with $\beta=10$ within each cluster; While in Skew, each client equally possesses data shards with 5 classes. Our iPFL achieves comparable or even better model performance and the highest utility across 9 settings.}
    \label{fig:main_scatter}
\end{figure}

\subsection{Performance Evaluation} \label{ch:results}
We use five image and text classification datasets commonly used in federated learning literature; and four instruction-tuning datasets for training instruction-following large language models.
Classification datasets includes CIFAR-10~\cite{cifar10}, Fashion-MNIST~\cite{fashion}, PACS~\cite{pacs}, FEMNIST~\cite{leaf}, and Shakespeare~\cite{leaf};
while instruction-tuning datasets includes three financial datasets (FIQA~\cite{Maia2018WWW18OC}, TFNS~\cite{tfns2022}, and NWGI~\cite{fingpt_github}) and a coding dataset~\cite{codealpaca}.
We compare our algorithm iPFL with other 7 baselines, including two general FL algorithms--FedAvg~\cite{fedavg} and FedProx~\cite{fedprox}, and 5 classical PFL algorithms--Ditto~\cite{ditto}, FedAMP~\cite{fedamp}, CFL~\cite{cfl}, FedFomo~\cite{fedfomo} and pFedGraph~\cite{pfedgraph}.

To evaluate the economic performance of iPFL, we introduce the utility function, as defined in~\cref{eq:utility}. It consists of three components: collaboration gain (\cref{eq:G}) with preference $K$, the sharing cost with individual unwillingness $c$ and accumulated payment (\cref{eq:p}) in all rounds. To evaluate model performance, we utilize the classification accuracy metric in classification tasks. For evaluation in instruction-tuning tasks, we utilize the corresponding test dataset for financial clients to evaluate accuracy and Humaneval~\cite{humaneval} for coding clients to evaluate passing rate.
The baselines, specific settings for $K$ and $c$, and implementation details are provided in the experimental section in supplementary information.

For classification tasks, we consider 9 settings with 5 datasets. For CIFAR-10 and Fashion-MNIST, we design three types of data partition among clients: termed as \textit{NIID}, \textit{Cluster}, and \textit{Skew}. 
(i) NIID is a common setting~\cite{fedma,bayesian,feddyn,fedcog}, where local data among clients follows the Dirichlet distribution (default $\beta=0.1$). 
% A smaller $\beta$ implies more severe data heterogeneity among clients.
(ii) The Cluster involves random client clustering, distinguishing between high (smaller $\beta$) and low heterogeneous levels within and between groups.
(iii) For Skew, total classes are divided into clusters so that in each cluster, each client possesses 5 classes.
FEMNIST and Shakespeare exhibit natural heterogeneity. In the case of PACS with four domains, each cluster represents one domain, namely the Cluster partition.
The evaluation results of our iPFL against eight baselines across nine settings are shown in \cref{fig:main_scatter}, emphasizing the comparisons on the trade-off between model performance and economic utility.
Our iPFL achieves a comparable or even better performance with performance-oriented baselines. 
Meanwhile, iPFL consistently excels in economic utility, as evidenced by its highest plot scatter across diverse settings.
Specifically, iPFL outperforms FedAMP by 1.75\% in accuracy and 217.4 in utility, respectively. 
% \Note{yr: fill the xx. suggest Ditto. }
Overall, these results show that iPFL effectively strikes a balance between model performance and economic utility, demonstrating its capacity to harmonize model performance and economic benefits within the personalized federated learning framework.

For the instruction-tuning tasks, we consider two scenarios.
(i) We configure a scenario for financial sentiment analysis with six clients, where every two clients share one of the following datasets: FIQA~\cite{Maia2018WWW18OC}, TFNS~\cite{tfns2022}, or NWGI~\cite{fingpt_github}.
(ii) We consider a more complex scenario to represent a higher heterogeneity level, where five clients possess the code data from CodeAlpaca~\cite{codealpaca} and three clients own the financial data from NWGI.
The results presented in~\cref{tab:results-financial} demonstrate the superiority of iPFL: it excels in both accuracy and utility metrics across scenarios.
For example, iPFL demonstrates a remarkable 5.55\% improvement in accuracy and a 58.3 gain in utility on the financial scenario compared to other baselines.
Besides, in the second setting, other baselines are inferior to local training in utility for some clients, failing to guarantee the IR property. 
% \Note{yr: cannot be seen from the table?}
This dual achievement highlights the effectiveness of our approach in enhancing model performance and economic utility.

\begin{table}[t]
    \centering
    \setlength\tabcolsep{4pt}
    \caption{Comparisons of model performance (accuracy or passing rate, \%) and utility of different algorithms on two instruction-tuning scenarios. The first scenario is the financial scenario, consisting of two clients for FIQA, TFNS, and NWGI, respectively. The second scenario involves three fiancial clients and five coding clients. Our iPFL consistently outperforms other baselines in both model performance and utility.}
    \label{tab:results-financial}
    \begin{tabular}{c|cccc|ccc}
    \toprule
    Scenarios & \multicolumn{4}{c|}{Finance} & \multicolumn{3}{c}{Finance \& Code} \\
    Evaluation & FIQA & TFNS & NWGI &  Avg-Utility &  NWGI & Code & Avg-Utility  \\
    \midrule
    Local & 84.02$\pm$6.09 & 80.58$\pm$0.83 & 43.17$\pm$4.48 & 0.0$\pm$0.0 & 50.61$\pm$2.63 & 13.54$\pm$2.30 & 0.0$\pm$0.0 \\
    \hline
    FedAvg & 78.19$\pm$1.94 & 81.25$\pm$5.30 & 52.25$\pm$4.77 & 45.9$\pm$0.0 & 49.94$\pm$2.46 & 15.00$\pm$0.70 & 58.2$\pm$150.4  \\
    FedProx &78.55$\pm$0.40 & 80.56$\pm$6.28 & 52.44$\pm$1.68 & 45.9$\pm$0.0& 49.61$\pm$2.67 & 15.00$\pm$1.26 & 58.2$\pm$150.4\\
    FedAMP & 84.01$\pm$4.03 & 76.63$\pm$5.48 & 42.56$\pm$6.98 & 45.9$\pm$0.0 & 51.58$\pm$1.38 & 14.02$\pm$0.86 & 58.2$\pm$150.4 \\
    CFL & 85.11$\pm$6.61 & 77.06$\pm$6.98 & 45.94$\pm$4.68 & 45.9$\pm$0.0 & 52.28$\pm$4.97 & 14.15$\pm$0.80 & 58.2$\pm$150.4 \\
    pFedGraph & 83.65$\pm$5.57 & 76.75$\pm$5.66 & 47.94$\pm$3.09 & 45.9$\pm$0.0 & 50.00$\pm$5.65 & 14.27$\pm$1.76 & 137.7$\pm$291.6\\
    \hline
    \textbf{iPFL} & \textbf{85.47$\pm$6.10} & \textbf{83.38$\pm$2.83} & \textbf{56.25$\pm$1.06} & \textbf{96.5$\pm$0.0} & \textbf{53.11$\pm$0.51} & \textbf{15.85$\pm$1.14} & \textbf{208.1$\pm$131.1} \\
    \bottomrule
    \end{tabular}
\end{table}
\subsection{Incentive Properties}
\begin{figure}
    \centering
    \subfigure[CIFAR-10-Cluster]{
        \includegraphics[width = 0.32\textwidth]{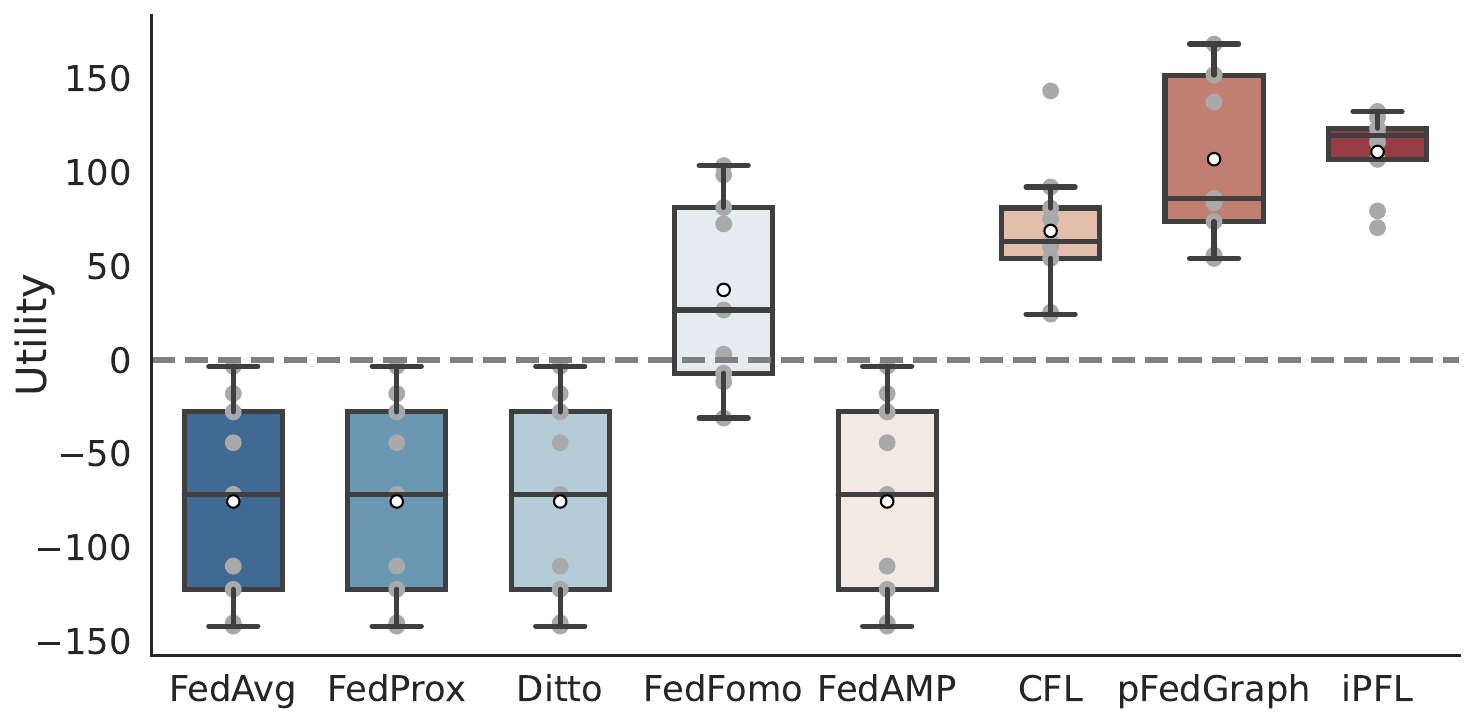}}
    \subfigure[PACS]{
        \includegraphics[width = 0.32\textwidth]{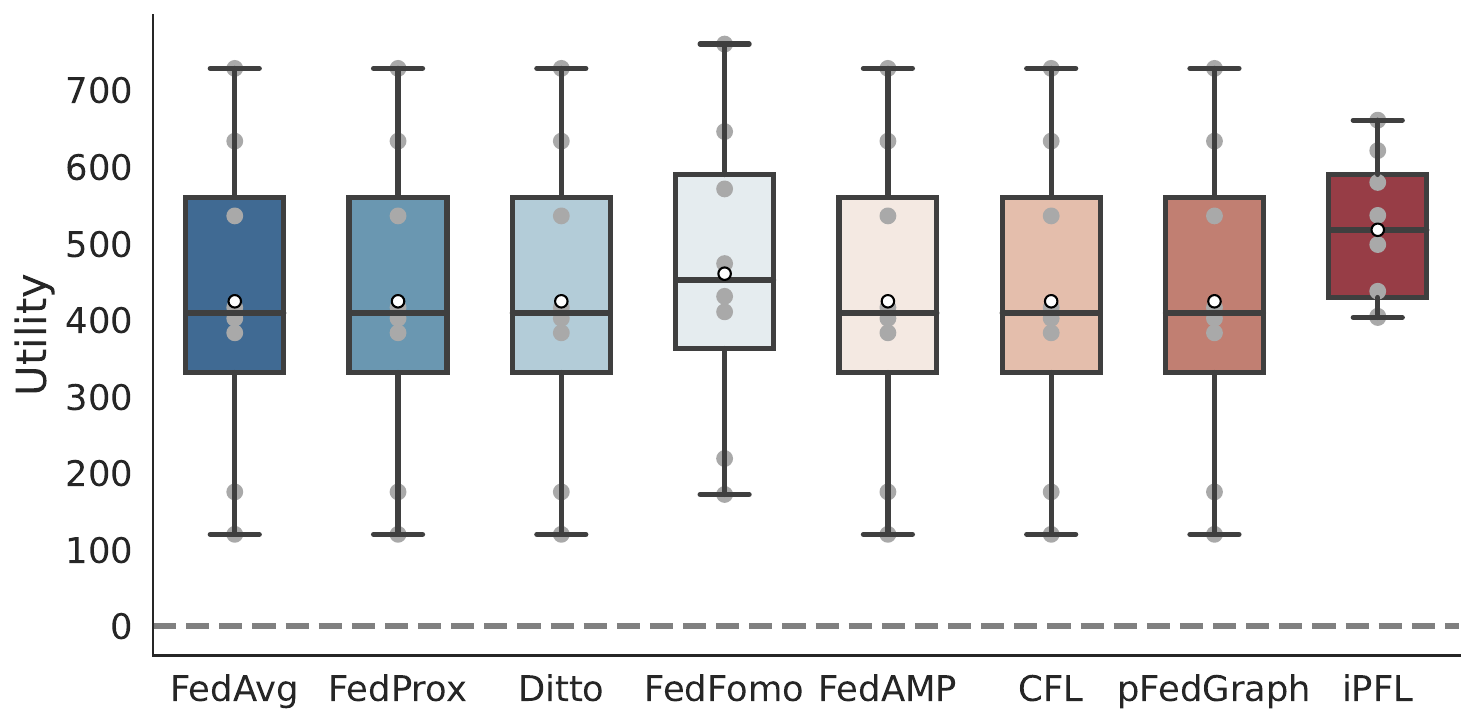}}
    \subfigure[Fashion-MNIST-NIID]{
        \includegraphics[width = 0.32\textwidth]{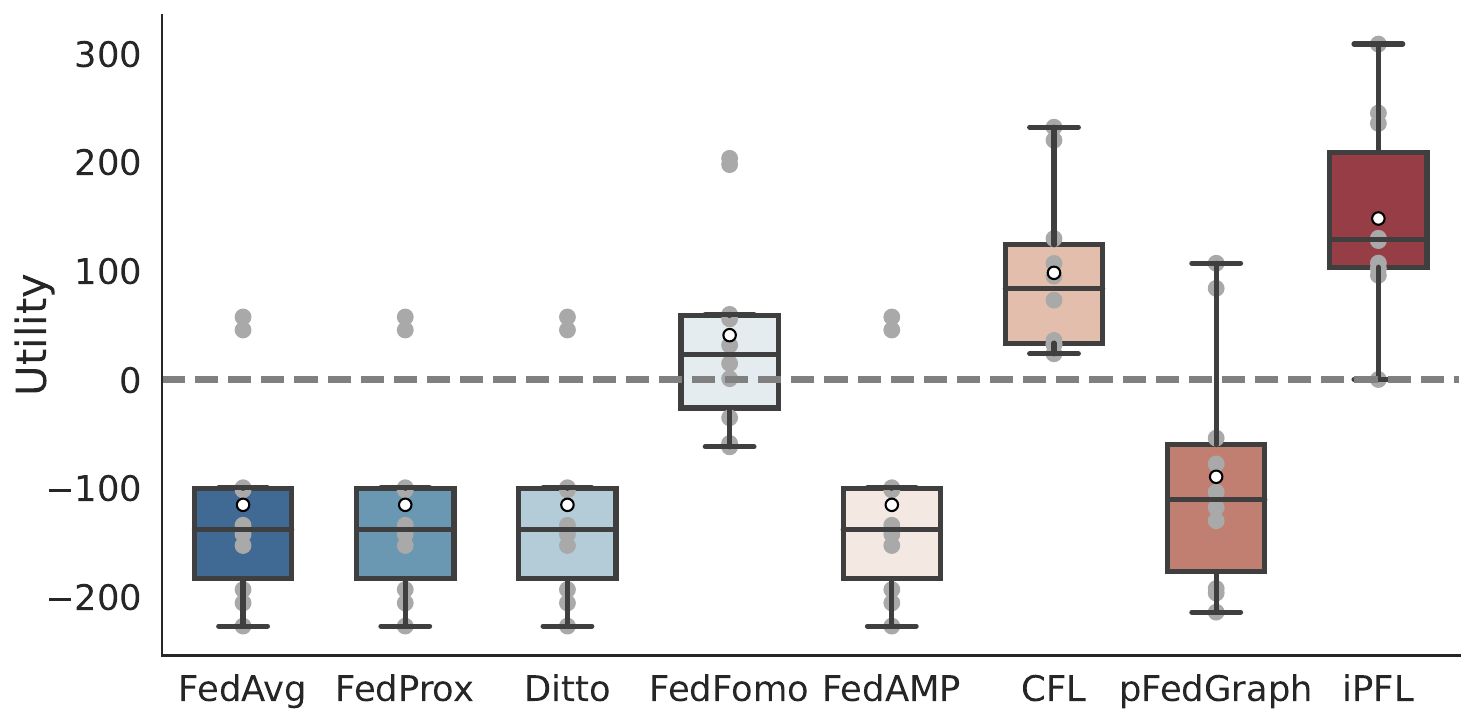}}
    \caption{The utility distribution of clients with different algorithms under three settings. Specifically, the circle denotes the mean utility of all clients, and the gray scatter represents the individual client utility values. Our iPFL guarantees positive utility for each client and achieves the highest average utility.}
    \label{fig:box_IR}
\end{figure}

\textbf{Individual rationality.}
Here, we show the individual client utility distribution on CIFAR-10-Cluster, PACS and Fashion-MNIST-NIID scenarios in~\cref{fig:box_IR}.
We compare iPFL with 7 representative baselines.
Remarkably, in these scenarios, our proposed iPFL ensures that the utility of each client remains positive, outperforming all the other algorithms.
These experiments convincingly verify that our proposed iPFL ensures the property of individual rationality (i.e., every participant benefits from the system), a critical property to incentivize institutions to join the market willingly~\cite{kang2019incentive, zeng2021comprehensive}.
Note that we accordingly provide the theoretical guarantee in~\cref{theorem_ir}.

\begin{figure}[t]
    \centering
    \includegraphics[width = 0.98\textwidth]{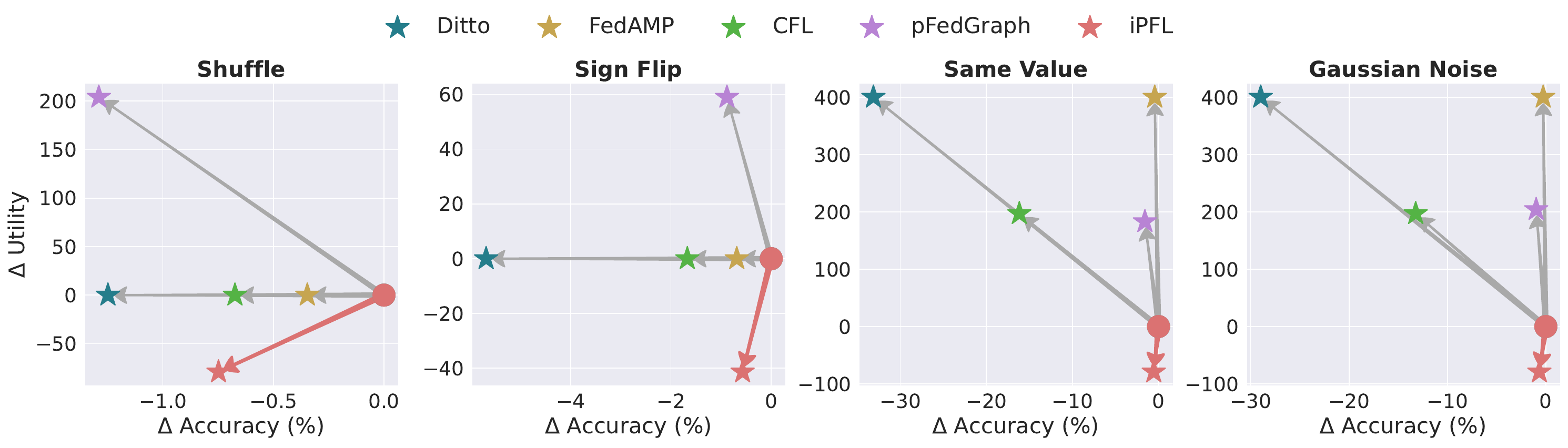}
    \caption{The change of average benign clients' performance (\%) and malicious client utility after 4 different attack types of an attacker, under the Cluster setting on CIFAR-10. For each algorithm, we utilize the circle $\medbullet$ and star $\bigstar$ to separately represent the benign clients' states with their mean accuracy and utility.}
    \label{fig:attack}
\end{figure}

\textbf{Robustness.}
In this part, we investigate the robustness of FL algorithms against four distinct types of model poisoning attackers~\cite{pfedgraph}.
Attack strategies include (a) shuffling model updates, (b) flipping the numerical sign of model updates, (c) manipulating model updates with the same value at each element, and (d) manipulating model updates based on random Gaussian noises. 
Based on the CIFAR-10-Cluster scenario, we conduct one experiment for each attack type, where one attacker is introduced.
We compare our iPFL with four representative state-of-the-art PFL algorithms: Ditto~\cite{ditto}, pFedGraph~\cite{pfedgraph}, CFL~\cite{cfl}, and FedAMP~\cite{fedamp}.
In~\cref{fig:attack}, we illustrate the changes in the averaged performance of benign clients and the utility of malicious clients after being exposed to attack.
Notably, only our iPFL succeeds in reducing the utility of the malicious attacker while simultaneously maintaining accuracy for the benign clients.
This unique characteristic positions our iPFL as a robust technical foundation for a healthy model-sharing market.

\textbf{Truthfulness.} 
Here, we explore the effects of clients being dishonest by considering a scenario where one client lies about the dataset size or training cost.
In \cref{tab:liars}, we show the liar's accuracy and utility over different lying ratios (compared with true value).
The table shows that in our iPFL, the liar always achieves lower or the same accuracy, and significantly lower utility.
These results verify that one cannot benefit by lying, which demonstrates the effectiveness of our iPFL in discouraging dishonest behaviors, contributing to promote the healthy development of the market.
Note that we accordingly provide the theoretical interpretation in \cref{theorem_truth}.

\begin{table}
    \centering
    \setlength\tabcolsep{8pt}
    \caption{Liar' performance(\%) and utility comparison under different lying ratios (1 denotes honest) of her true data size or cost, under the NIID setting of CIFAR-10. Lying on reported private information causes performance degradation and loss of earnings.}
    \label{tab:liars}
    \begin{tabular}{c|c|ccc|ccc}
    \toprule
    Cases & Honest & \multicolumn{3}{c|}{Lying on data size} & \multicolumn{3}{c}{Lying on cost} \\
    \midrule
    Lying Ratio & 1 & 0.1 & 0.5 & 10 & 2 & 5 & 10 \\
    Liar's Accuracy & \textbf{75.430} & 66.933 & 75.331 & 75.430 & 75.430 & 75.430 & 75.430 \\
    Liar's Utility & \textbf{617.295} & -513.410 & -9.951 & 0.00 & 0.00 & 0.00 & 0.00 \\
    \bottomrule
\end{tabular}
\end{table}

\subsection{Inclusive Market}
To verify that our iPFL is inclusive that can include clients with diverse preferences and economic conditions, we simulate a market that consists clients with diverse roles.
In the market, some traders buy and sell models, buyers who only buy models, sellers who only sell models, and attackers who try to sell poisoned models (we use a randomly parameterized model).
These are achieved by setting the profiles of clients: we set the level of data eagerness as a random positive value for traders and buyers, while zero for sellers and attackers; we set the cost as a random positive number for traders, $+\infty$ for buyers, zero for sellers and attackers; see details in supplementary material.
Finally, we build a market based on CIFAR-10-Cluster scenario with 12 clients and conduct model training for 20 rounds.
We record the accumulated money transaction and the accuracy difference between model trained by iPFL and local training, and demonstrate them in~\cref{fig:playground}.
From the figure, we can clearly see that the transactions among the clients are well aligned with their roles (i.e., purposes).
The traders buy models from others to obtain models with higher accuracy, and sell models to others to make a profit at the same time.
The buyers pay others to buy models to improve their models while the sellers earn money by selling models.
The attacker is successfully isolated by others, doing no harm to the market.
Overall, the experiments verify that iPFL is an incentivized and inclusive PFL system since every unique individual gains benefits from joining the system. 

\section{Discussion}
In response to the challenges posed by the depletion of publicly available data and the need for collaboration among private institutions, we establish an inclusive sharing market that incentivizes the contributions of diverse participants with unique model preferences and economic conditions.
Rooted in the personalized federated learning paradigm, iPFL integrates a graphical game within the framework based on the directed collaboration graph. 
iPFL introduces a novel and multifaceted objective, aiming to minimize loss on relevant tasks, maximize pairwise model similarity, and enhance overall social welfare within the system. 
Our iPFL framework modifies local training methods to achieve improved personalization, flexibly adjusts collaboration regarding models and economic conditions, and implements a sophisticated payment mechanism.
The proposed system iPFL facilitates training, collaboration, and transactions to meet each participant's demands and achieves incentive properties theoretically and experimentally. Regarding privacy preservation, our iPFL avoids direct data sharing, ensuring effective data isolation.
% maintains privacy levels equivalent to FedAvg
Regarding the communication overhead, participants in iPFL only need to additionally report their model preference $K_i$, cost $c_i$ and data amount $N_i$ at the start of training. These one-time uploads are negligible for communication but significantly improve the ability to balance model performance and economic utility. 

\begin{figure}
    \centering
    \includegraphics[width=1\linewidth]{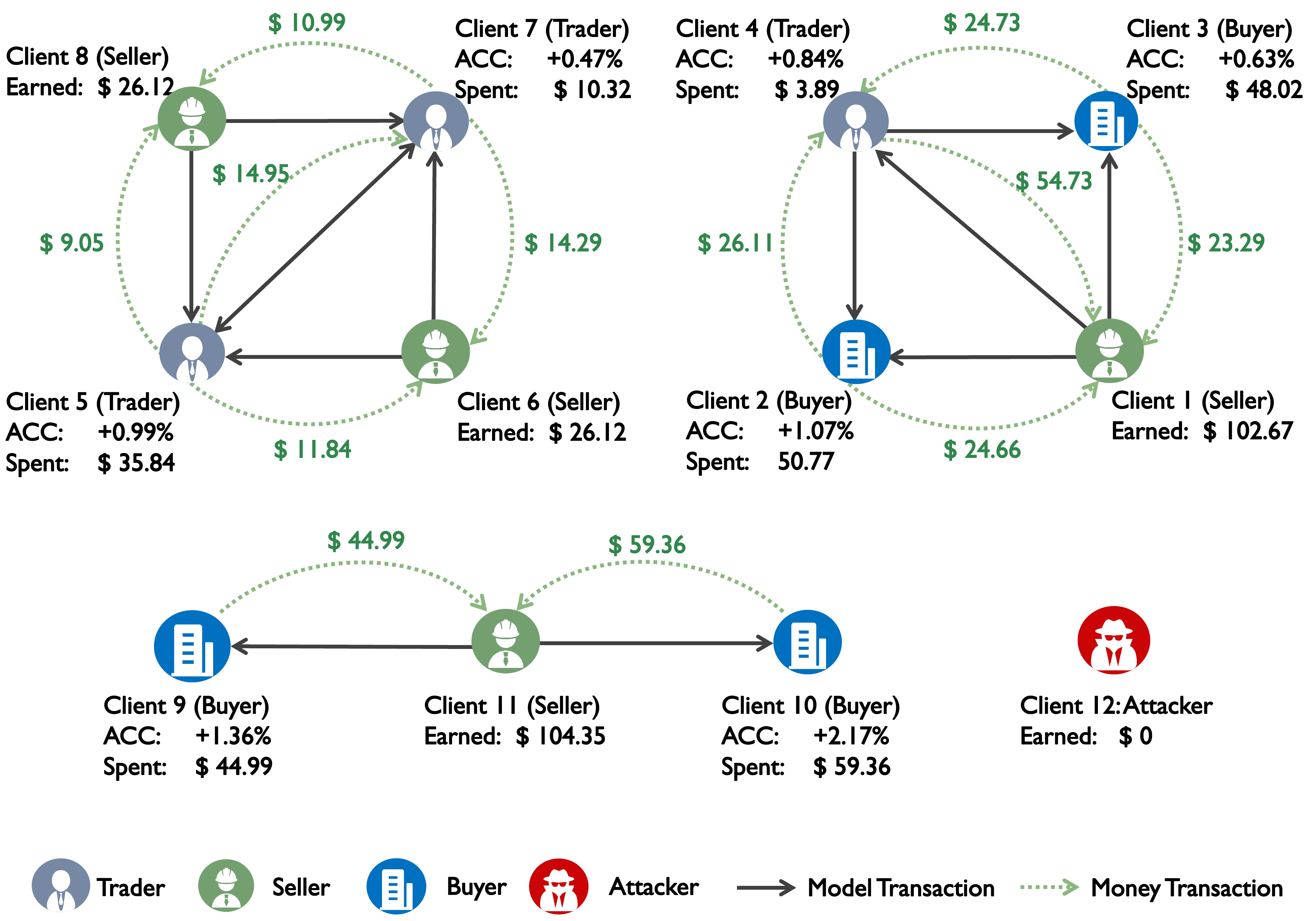}
    \caption{The transaction graph of market simulation.}
    \label{fig:playground}
\end{figure}

Comprehensive experiments reveal several significant findings about our iPFL.
First, iPFL demonstrates exceptional versatility in balancing model performance and economic utility of the AI landscape. Extensive experiments, spanning various machine learning tasks and model scales in~\cref{fig:main_scatter} and~\cref{tab:results-financial}, highlight its capability to achieve comparable or superior model performance and consistently highest social welfare.
Second, iPFL ensures individual rationality,  as every institution involved in the system achieves non-negative benefits (see~\cref{fig:box_IR}). This inherent motivation acts as a catalyst, encouraging a growing number of institutions to join the ecosystem. This, in turn, leads to an expansion of the market size, fostering a resilient and extensive database that can further catalyze advancements in AI research.
Third, iPFL exhibits a remarkable capability to prevent dishonest practices. Exaggerating data size and cost by participants results in reduced utility (shown in~\cref{tab:liars}), acting as a deterrent against dishonest behavior and market fraud. This feature underscores iPFL's commitment to fostering an environment of honesty and integrity.
Fourth, iPFL showcases a robust defense against potential attackers. Achieved by effectively isolating malicious participants in~\cref{fig:attack}, iPFL contributes to a stable and trustworthy market environment.
In addition, our inclusive simulation experiment in~\cref{fig:playground} further supports these findings. It demonstrates that honest institutions with distinct needs can acquire what they require, showcasing iPFL as an epitome of an actual healthy market. 

Through these advancements, iPFL paves the way for a new era in collaborative AI. With iPFL, institutions can not only benefit from personalized models but also actively contribute to and gain from a flourishing inclusive market while preserving privacy. However, our work also has limitations.  Our work assumes the static nature of data, economic needs, and participants' willingness to join during the entire training process. However, in practical scenarios, institutions may choose to exit the training process. For instance, buyers may not require the model for specific tasks or seek more attractive markets. Our framework, designed under the assumption of a static federation, may not fully accommodate such dynamic transformations, especially autonomous exits or joins. 
Future research could explore more flexible frameworks that adapt to the dynamic states, by adjusting model-sharing strategies, or pricing mechanisms.

\section{Methods}

In this section, we present the introduction of our method, iPFL.
We first provide some key definitions and formulate the objective of our method.
Then, we introduce the details of our method, including training procedure and payment.
Finally, we provide theoretical insights to discuss about the incentive properties and robustness of our method iPFL.

\subsection{Problem Formulation and Training Objective}

We consider the popular PFL settings, where $m$ institutions join the system as clients and are managed by a central server. Each client $i$ holds a private dataset $\mathbf{Z}_i=(\mathbf{z}_{i,1},...,\mathbf{z}_{i,N_i})$ with $N_i$ data points sampled from client $i$'s local data distribution $\mathcal{D}_i$. Each client $i$ maintains its own model parameters $\theta_i$. Given a common loss criterion $l(\cdot,\cdot)$, the empirical loss of client $i$ on its own dataset $\mathbf{Z}_i$ is: $L_i(\theta_i)=\frac{1}{N_i}\sum_{k\in[N_i]}l(\theta_i;\mathbf{z}_{i,k})$. The clients hope to train a personalized model that performs well on its local data distribution $\mathcal{D}_i$. In this case, the population loss (testing loss) for client $i$ is $L^*_i(\theta_i)=\mathbb{E}_{\mathbf{z}\sim\mathcal{D}_i}l(\theta_i;\mathbf{z})$.

Due to privacy concerns and communication constraints in multi-institutional scenarios, the clients cannot directly send their data to other clients. In our work, we consider the model-sharing strategy in PFL. As a member of the federation, each client can refer to others' model parameters, coordinated by the server and realized at the server side. Specifically, to describe the model sharing to topology among the clients, we use a directed graph represented by the adjacency matrix $\mathbf{A}=(a_{ij})_{m\times m}$ with:
$$
a_{ij}=
\begin{cases}
1&\text{if $i\neq j$ and client $i$ imports the model of client $j$}\\
0&\text{else}
\end{cases}
$$

Each client may participate in federated learning with different purposes (i.e., for a better model for their local tasks or economic rewards). Meanwhile, we assume that the server has no interest in clients' tasks and only acts as a neutral coordinator. With the model sharing topology $\mathbf{A}$, we then use a graphical game model to formulate the market of PFL. The players in this game are the clients, without the server. In the market, the clients can choose the models they will import, so the action of client $i$ is described by $\mathbf{a}_i$ (the $i$th roll of $\mathbf{A}$). The action of model sharing can reflect the fairness of the training. For example, a client may share its own model with many other clients, but import few models from them. In this case, this client may not be satisfied with the arrangement of training, as it cannot obtain proportionate treatment from the federation. Therefore, we additionally introduce a utility function to capture the clients' non-training benefits (i.e., economic gain, satisfaction with the collaboration) in the procedure of training. 

\vspace{3mm}\begin{definition}[utility]
    Consider that the clients are sharing models for multiple rounds (in each round $t$ the clients share models according to $\mathbf{A}^t$). The utility of client $i$ in each round is defined as:
    \begin{align}
        U^t_i=G_i(\mathbf{a}^t_i)-\sum_{j\in[m]}{a^t_{ji}c_i}-p^t_i.
    \end{align}
    \label{eq:utility}
\end{definition}\vspace{3mm}

We elaborate on the three components as follows:

\vspace{3mm}\begin{definition}[collaboration gain]
    $G_i:\{0,1\}^m\to\mathbb{R}$ is the gain function for client $i$, with $G_i(\mathbf{a}^t_i)$ describing client $i$'s gain of data resource from chosen collaborators $\mathbf{a}^t_i$.
\end{definition}\vspace{3mm}

In our work, we consider a specified $G_i$ to describe the diminishing collaboration gain for the quality-aware clients. We assume that for client $i$ the gain is related to the amount of imported data, which can be represented by a continuous and concave function $g_i:\mathbb{R}\to\mathbb{R}$. Specifically, we consider:
\begin{align}
      G_i(\mathbf{a}_i)=g_i(\sum_{j\in[m]}a_{ij}{N_j})=\sqrt{\frac{K_i}{N_i}}-\sqrt{\frac{K_i}{N_i+\sum_{j\in[m]}a_{ij}{N_j}}},
      \label{eq:G}
\end{align}
with hyper-parameter $K_i$ representing client $i$'s level of eagerness for data. $K_i$ can be varied among the clients , reflecting their different data needs. If $K_i=0$, that means the client cannot benefit from enlarged data access, so he may be a pure model seller and will not buy models from any other client. If $K_i>0$, getting more data from the collaborators will increase the gain, but the marginal benefit brought by each collaborator will become smaller. If there is no collaborator, the gain is zero, as $G_i(\mathbf{0})=g_i(0)=0$.

\vspace{3mm}\begin{definition}[sharing cost]
    If $\theta^t_i$ is imported by one another client, client $i$ will suffer a loss of $c_i$. 
\end{definition}\vspace{3mm}

$c_i$ indicates client $i$'s unwillingness to share its model due to fairness or privacy concerns. By definition, unless the payback is larger than $c_i$, the change of utility is negative and client $i$ is reluctant to share its model. From this perspective, $c_i$ can also be taken as a minimum price to share the model. So we introduce money transactions to overcome this barrier to collaboration.

\vspace{3mm}\begin{definition}[overall payment]
    $p^t_i$ represents the amount of payment client $i$ should pay to the system in round $t$. 
\end{definition}\vspace{3mm}

If $p^t_i>0$, that means client $i$ has to pay for the benefits gained from the federation; otherwise, that means client $i$ is rewarded for his contribution to the federation. Since the server is neutral, we only consider the monetary transaction among the clients. Denote $r^t_{ij}$ as the remittance from client $i$ to client $j$. Since the money transaction is symmetric: $p^t_i=\sum_{j\in [m]}r^t_{ij}-\sum_{j\in [m]}r^t_{ji}$, there is $\sum_{i\in[m]}p^t_i=0$. 

Based on our evaluation of clients' utility, we have the social welfare of the whole market system.

\vspace{3mm}\begin{definition}[social welfare]
    Denote $\mathbf{c}=(c_1,...,c_m)^\top$, the \textbf{social welfare} at round $t$ is defined as:
    \begin{align}
        {\rm SW}(\mathbf{A}^t)=\sum_{i\in[m]}U_i^t=\sum_{i\in[m]}\bigg[G_i(\mathbf{a}^t_i)-\mathbf{c}^\top\mathbf{a}^t_i\bigg].
    \end{align}
    \label{eq:SW}
\end{definition}\vspace{3mm}

In iPFL, our goal is to build up an inclusive PFL system that provides personalized training according to clients' models and economic needs. To achieve this, we propose a novel PFL optimization problem, which pursues smaller training loss and larger between-collaborators model similarity, while maintaining the economic utility of all the clients. Define the similarity between two models $\theta_i$ and $\theta_j$ using a differentiable function $d(\theta_i,\theta_j)$. To enable more collaboration among more similar clients, we can include pair-wise similarity in the optimization problem of PFL. Thus, our training objective is defined as:
\begin{align}
   \min_{\forall i\in[m]:\theta_i;\mathbf{A}}\
   \sum_{i\in[m]}L_i(\theta_i)
   +\lambda\sum_{i,j\in[m]}a_{ij}\frac{N_j}{N_i}d(\theta_i,\theta_j)
   -\text{SW}(\mathbf{A}).
\label{eq:overall objective}
\end{align}
The first two terms $L_i(\theta_i)+\lambda\sum_{i,j\in[m]}a_{ij}\frac{N_j}{N_i}d(\theta_i,\theta_j)$ are model-similarity-aware training loss. With pairwise collaboration indicated by binary indicator $a_{ij}$, if client $i$ imports $j$'s model, a regularization term $\lambda\frac{N_j}{N_i}d(\theta_i,\theta_j)$ will be added to the training loss. The third term $\text{SW}(\mathbf{A})$ is the social welfare under the graph $\mathbf{A}$, which can also be taken as a regularization term to avoid $\mathbf{A}$ degrade to $\mathbf{0}$. Therefore, the clients can attain personalized models without losing generality by minimizing both the loss on local tasks and the model difference compared to their collaborators. At the same time, our system also optimizes social welfare by refining the clients' selection of references, which ensures the benefits of each client and makes the training more economic-efficient.

\subsection{Training Procedure}

\begin{algorithm}
\caption{Overview of iPFL}\label{alg:overview}
\begin{algorithmic}
    \State\textbf{input:} $m$ clients, each with a local private dataset for training.
    \State Server sends an initial model $\theta^{0}$ to every client.
    \For{$i=1$ to $m$}
        \State Client $i$ performs local training and obtains $\theta_i^1=\arg\min_{\theta_i}L_i(\theta_i)+\frac{\lambda}{2\eta}\|\theta_i^t-\bar\theta_i^{0}\|_2^2$.
        \State Client $i$ reports $N_i,c_i, G_i(\cdot)$ back to Server.
    \EndFor
    \For{$t=1$ to $T-1$}
        \For{$i=1$ to $m$}
            \State Client $i$ reports $\theta_i^t$ back to Server.
        \EndFor
        \For{$i=1$ to $m$}
            \State Server calculates $\mathbf{a}_i^t$ by Algorithm~\ref{alg:thresh}. \hfill \textcolor{blue!50}{\textbackslash\textbackslash~Graph Topology Learning}
        \EndFor
        \For{$i=1$ to $m$}
            \State Server calculates $p_i^t$ according to $\mathbf{A}^t$. \hfill \textcolor{blue!50}{\textbackslash\textbackslash~Payment Calculation}
            \If{Client $i$ pays $p^t_i$ to Server}
                \State Server calculates the prox-center $\bar\theta_i^t=\theta_i^{t}-\frac{\eta}{N_i}\sum_{j\in[m]}a_{ij}N_j\nabla_{\theta_i}d(\theta_i,\theta_j^{t})$.
                \State Server sends the prox-center model $\bar\theta_i^t$ to Client $i$.
                \State Client $i$ updates $\theta^{t+1}_i=\arg\min_{\theta_i}L_i(\theta_i)+\frac{\lambda}{2\eta}\|\theta_i^t-\bar\theta_i^{t}\|_2^2$. \hfill \textcolor{blue!50}{\textbackslash\textbackslash~Local Model Training}
            \Else
                \State Client $i$ quits and takes the best $\theta_i\in\{\theta_i^{t'}|t'\le t\}$.
            \EndIf
        \EndFor
    \EndFor
    \State\textbf{output:} $\theta_i,i\in[m]$ for each client.
\end{algorithmic}
\end{algorithm}

We overview the system in \cref{alg:overview}, which takes $T$ rounds in total to alternatively optimize personalized model $\theta_i$ and neighbor selection $\mathbf{a}_i$ for each client and assign appropriate payment among clients. Specifically, in each round $t$, the clients first update and upload their local models $\theta_{i}^t$. Then, the server calculates the data sharing graph $\mathbf{A}^t$ by optimizing the clients' actions according to the game model. The amount of payment $p_i^t$ for each client is also calculated according to $\mathbf{A}^t$ of round $t$. At the end of each round, the clients have two choices: 
1) pay $p_i^t$ and receive the aggregated model $\bar\theta_i^t$ for next round; 
2) quit the federation and take the best model in previous rounds as the final model.
Therefore, the training procedure involves three key steps: local model training at the client side, where personalized models are trained locally; graph topology learning at the server side, where the model sharing topology is learnt from the uploaded local models; payment calculation at the server side, where a bill for each client is determined by the server and the clients complete money transaction by paying the bill.

\textbf{Local model training}. To train a personalized model by collective data, each client updates its model parameters locally by simultaneously minimizing loss on local tasks and model-level distance from the selected collaborators' models. Since it is not feasible to optimize all the clients' models at the same time, we update the local models by block gradient descent. That is, in round $t$, each client updates its model by:
\begin{align}
    \theta_i^{t+1}
    =\arg\min_{\theta_i}L_i(\theta_i)+\lambda\sum_{j\in[m]}a_{ij}^t\frac{N_j}{N_i}d(\theta_i,\theta_j^{t}),
\label{eq:local_bcd_obj}
\end{align}
where $a^t_{ij}$  is the collaboration indicator in $\mathbf{A}^t$ determined by the server at round $t$. If the server determines that client $i$ should collaborate with client $j$, $a_{ij}$ will be set as $1$, so that client $i$ will be encouraged to learn from client $j$ during the local model training by minimizing the distance between local model $\theta_i$ and collaborators' models $\{ \theta_j^t \}_{a_{ij}=1}$. However, directly solving \cref{eq:local_bcd_obj} requires times of communication cost because client $i$ need access to all collaborators' local models. To efficiently avoid introducing additional communication costs, we propose to apply the proximal gradient descent method, in which the server computes \cref{eq:prox_step} in advance before transmitting information to clients and each client optimizes \cref{eq:local_obj} during local model training in practical implementation:
\begin{align}
    &\bar\theta_i^{t}
    =\theta_i^{t}-\frac\eta{N_i}\sum_{j\in[m]}a_{ij}^tN_j\nabla_{\theta^{t}_i}d(\theta^{t}_i,\theta_j^{t})
\label{eq:prox_step}\\
    &\theta_i^{t+1}
    =\arg\min_{\theta_i}L_i(\theta_i)+\frac{\lambda}{2\eta}\|\theta_i -  \bar\theta_i^{t}\|_2^2
\label{eq:local_obj},
\end{align}
where $\eta$ is the step size in the calculation of the proximal center. With such a technique, the server only needs to send client $i$ a proximal center $\bar\theta_i^{t}$ at round $t$ instead of multiple models from collaborators.

\textbf{Graph topology learning}. The server needs to find a suitable model-sharing graph based on the local models uploaded by clients. The graph is optimized by minimizing the model distance between collaborators and maximizing social welfare of the overall system, which corresponds to solving a sub-problem of \cref{eq:overall objective}:
\begin{align}
\forall i\in[m]:\mathbf{a}_i^t=\arg\min_{\mathbf{a}_i}\phi_i(\mathbf{a}_i)=\lambda\sum_{j\in[m]}a_{ij}\frac{N_j}{N_i}d(\theta^t_i,\theta^t_j)+\mathbf{c^\top a}_{i}-G_i(\mathbf{a}_i)
\label{eq:gl_obj_rewrite}
\end{align}
$$
\text{s.t. }\forall i,j\in[m]:a_{ij}\in\{0,1\}.
$$
where $\phi_i$ denotes the objective function of the sub-problem for each client. The problem is an NP-hard integer programming and finding an optimal solution can be very costly. Therefore, we propose an efficient graph learning algorithm (\cref{alg:thresh}) to get an approximate solution for this optimization problem in $O(m)$ time. In this algorithm, we calculate a threshold data amount $N^{Th}_k$ for each potential collaborator of client $i$ by $g_i(N^{Th}_j)-g_i(N^{Th}_j-N_j)=c_j+\lambda\frac{N_j}{N_i}d(\theta_i^t,\theta_j^t)$. Since the marginal collaboration gain brought by each collaborator of client $i$ decreases with client $i$'s total accessible data amount, if $a_{ij}=0$ and $N^{Th}_j>N_j+\sum_{k\in[m]}a_{ik}{N_k}$, then setting $a_{ij}=1$ would make $\phi_i$ smaller. Therefore, \cref{alg:thresh} keeps adding the client $j$ with the largest $N^{Th}_j$ to the collaborators of client $i$ until $\forall j:a_{ij}=0\Rightarrow N^{Th}_j\le N_j+\sum_{k\in[m]}a_{ik}{N_k}$. Hence, after \cref{alg:thresh} reaches the condition of termination, we have a solution $\mathbf{a}^*_i$ that satisfies:
\begin{align}
    &\forall{j\neq i}:a^*_{ij}=1\Rightarrow\phi_i(\mathbf{a}^*_i-\mathbf{e}_j)
            >\phi_i(\mathbf{a}^*_i)\\
    &\forall{j\neq i}:a^*_{ij}=0\Rightarrow\phi_i(\mathbf{a}^*_i+\mathbf{e}_j)
            \ge\phi_i(\mathbf{a}^*_i),\notag
\end{align}
where $\mathbf{e}_j$ is the $j$th row of the identity matrix. Though this algorithm cannot ensure a globally optimal solution to \cref{eq:gl_obj_rewrite}, its solution $\mathbf{a}^*_i$ is a locally optimal choice for client $i$, as adding or removing any collaborator will not make the objective $\phi_i$ smaller. By introducing such an approximate solution, our graph learning algorithm can attain a feasible $\mathbf{A}^t$ efficiently and robustly without sacrificing a large amount of time searching for unnecessary optimality, which is sufficiently effective in practice.

\begin{algorithm}[t]
  \caption{Graph Topology Learning}\label{alg:thresh}
  \begin{algorithmic}
    \State\textbf{input:} $g_i,\{\theta_1^t,...,\theta_m^t\},\{c_1,... ,c_m\},\{N_1,... ,N_m\}$. \hfill \textcolor{blue!50}{\textbackslash\textbackslash~Here $g_i(x)=\sqrt{\frac{K_i}{N_i}}-\sqrt{\frac{K_i}{N_i+x}}$}
        \State\textbf{initialization:} $\mathbf{a}_i=\mathbf{0},n=0$
        \For{$j=1$ to $m$}
            \State Calculate threshold $N^{Th}_j$ by solving $g_i(N^{Th}_j)-g_i(N^{Th}_j-N_j)=c_j+\lambda\frac{N_j}{N_i}d(\theta_i^t,\theta_j^t)$.
            \If{no solution}
                \State Set $N^{Th}_j=0$.
            \EndIf
        \EndFor
        \For{$j=1$ to $m$}
            \State $k = \arg\max_j N^{Th}_j\quad s.t. a_{ij}=0$
            \If{$n+N_k<N^{Th}_k$}
                \State $a_{ik}=1$ \hfill \textcolor{blue!50}{\textbackslash\textbackslash~Add the remaining client with the largest threshold}
                \State $n=n+N_k$
            \Else
                \State \textbf{break} \hfill \textcolor{blue!50}{\textbackslash\textbackslash~Stop adding if total data amount reaches the threshold}
            \EndIf
        \EndFor
    \State\textbf{output:} $\mathbf{a}_i$.
\end{algorithmic}
\end{algorithm}

\textbf{Payment calculation}. According to the definition of the utility of clients in \cref{eq:utility}, if client $i$ imports the model from $j$, it will pose a cost of $c_j$ to client $j$. This indicates that the model sharing is not reciprocal, leading to the dilemma that some clients lack the incentive to join the training. Therefore, after confirming the collaboration graph $\mathbf{A}^t$ among clients, the server needs to determine the required payment $p^t_i$ for client $i$, which needs to ensure that contributions from clients are aptly rewarded and those benefiting from these contributions are appropriately charged. In our payment design, we consider the reward calculated based on the benefit brought by the imported model. The payment is defined as follows: if client $i$ imports $j$'s model, client $i$ pays to $j$ the marginal benefit minus the model difference:
$$
    r^t_{ij}=a^t_{ij}[G_i(\mathbf{a}^t_i)-G_i(\mathbf{a}^t_i-\mathbf{e}_j)-\lambda\frac{N_j}{N_i}d(\theta^t_i,\theta^t_j)],
$$
where $G_i(\mathbf{a}^t_i)-G_i(\mathbf{a}^t_i-\mathbf{e}_j)$ is the marginal benefit (change of the gain) brought by $j$'s model and $\lambda\frac{N_j}{N_i}d(\theta^t_i,\theta^t_j)$ is the model difference term defined in \cref{eq:overall objective}. In this way, $p^t_i$ can be written as:
\begin{align}
\small
    p^t_i
    &=\sum_{j\in [m]}r^t_{ij}-\sum_{j\in [m]}r^t_{ji}\notag\\
    &=\sum_{j:a^t_{ij}=1}\bigg[G_i(\mathbf{a}^t_i)-G_i(\mathbf{a}^t_i-\mathbf{e}_j)
    -\lambda\frac{N_j}{N_i}d(\theta^t_i,\theta^t_j)\bigg]
    -\sum_{j:a^t_{ji}=1}\bigg[G_j(\mathbf{a}^t_j)-G_j(\mathbf{a}^t_j-\mathbf{e}_i)
    -\lambda\frac{N_i}{N_j}d(\theta^t_j,\theta^t_i)\bigg].
\label{eq:p}
\end{align}
We can see that this payment policy is beneficial for both client $i$ and $j$: while client $j$ gets paid more than minimal price, client $i$ does not lose all the benefits brought by client $j$'s model. Therefore, the model transaction in our system is reciprocal and no client is conveying benefits to others for free. Also, different from simply covering client $j$'s cost by setting $r^t_{ij}=c_j$, the clients cannot directly affect the payment by manipulating $c_i$. This can significantly reduce the regret of pricing (i.e., losing money for not setting $c_i$ higher) and greedy behaviors (i.e., reporting higher $c_i$ for more profits).

\subsection{Theoretical Insights}
\label{sec:inc_anly}
In this section, we provide some theoretical discussion to show the special properties of our system. To begin with, we show that our system is \textit{individual rational} and \textit{truthful}, which are two critical properties for an incentive mechanism. First,\cref{theorem_ir} ensures that the clients will be satisfied with the training arrangement (the proofs in this section are in Appendix).

\vspace{3mm}\begin{theorem}[Individual Rationality]
\label{theorem_ir}
If $\mathbf{A}^t$ is given by \cref{alg:thresh}, then $\forall{i\in[m],t\in[T]}:U_i^t=G_i(\mathbf{a}^t_i)-\sum_{j\in[m]}{a^t_{ji}c_i}-p^t_i\ge0$.
\end{theorem}\vspace{3mm}

Second, \cref{theorem_ic}  shows that the claim of $c_i$ is incentive compatible, as increasing $c_i$ will result in less sharing of client $i$'s model. So clients can control the spread of their models.

\vspace{3mm}\begin{lemma}[Incentive Compatibility of $c_i$]
\label{theorem_ic}
Denote $\mathbf{A}^t$ the graph calculated by the server when everyone honestly reports their $c_i$ and $\mathbf{\hat{A}}^t$ the new graph when client $i$ report $\hat{c}_i>c_i$. Then $\{j|\hat{a}^t_{ji}=1\}\subseteq\{j|a^t_{ji}=1\}$.
\end{lemma}\vspace{3mm}

On the basis of  \cref{theorem_ic} we can additionally prove \cref{theorem_truth}, which ensures that the clients cannot obtain additional income by overstating $c_i$.

\vspace{3mm}\begin{theorem}[Truthfulness]
\label{theorem_truth}
Denote $U_i$ the one-round utility of client $i$ when everyone honestly reports their $c_i$ and $\hat U_i^t$ as its utility when client $i$ reports $\hat c_i>c_i$. Then $\forall t:\hat U_i^t\le U_i^t$.
\end{theorem}\vspace{3mm}

At the same time, if client $i$ reports $\hat c_i<c_i$, it risks selling his model at a low price and the mechanism cannot ensure $U_i^t>0$. Thus, the clients are encouraged to reveal their true cost $c_i$ to the server. This contributes to harmonious collaboration because clients do not need to be secretive about their unwillingness to share.

Besides incentive properties, we then discuss our system's robustness against abnormally reported data amount $N_i$ and model parameters $\theta^t_i$. For benign and quality-aware clients, they have no reason to be dishonest about $N_i$ and $\theta^t_i$ as lying about $N_i$ and $\theta^t_i$ is harmful to their models: reporting wrong $N_i$ will result in inaccurate model aggregation; uploading fake $\theta^t_i$ may result in less personalization. However, malicious attackers can upload noisy models to attack the system or exaggerate their data amount to defraud extra payment and increase their weight in others' models. To address this issue, client selection procedure considers both data amount and model similarity in \cref{alg:thresh}. \cref{theorem_robust} shows that malicious clients who upload abnormal $N_i$ and $\theta_i$ are likely to be isolated without introducing extra efforts of model verification (e.g., testing models on a validation set).

\vspace{3mm}\begin{theorem}[Robustness against abnormal data amount]
\label{theorem_robust}
If $\mathbf{A}^t$ is given by \cref{alg:thresh} and $N_i\to+\infty$, then $\forall{j\in[m]}:a^t_{ji}=0$.
\end{theorem}\vspace{3mm}

As a result, the malicious clients whose priority is attacking the federation can only be trusted by other clients when they report a relatively smaller data volume. This means that their impact is limited: if client $j$ reports a smaller $N_j$, it will receive a smaller reference weight ($N_j/N_i$) in the second term of \cref{eq:local_bcd_obj} and other clients will not strongly emphasize its parameters in aggregation.

Besides, the consequence of uploading a fake model is similar. If client $i$ is malicious and uploads a fake model $\theta_i'$ (e.g., perturbing the real model parameters), $\theta_i'$ is very likely to be different from other clients' normal models that are trained on real datasets, indicating that the model difference term $\|\theta_i-\theta_j\|_2^2$ is large, which will increase the value of $N_i^{Thresh}$ calculated by other clients and make client $i$ less possible to be chosen. Therefore, the influence of such malicious clients is also limited.

\bibliographystyle{unsrt}  
\bibliography{references}  %%% Remove comment to use the external .bib file (using bibtex).


\end{document}